\documentclass[letterpaper, 10 pt, journal, twoside]{IEEEtran} 
\usepackage{graphics}    
\usepackage{times}       
\usepackage{amsmath}     
\usepackage{amssymb}     
\usepackage{graphicx}
\usepackage{subcaption}
\usepackage{url}
\usepackage{algorithm}
\usepackage[noend]{algpseudocode}
\usepackage{xcolor}
\usepackage{amsopn}

\newcommand{\bH}{\mathbf{H}}

\newcommand{\bX}{\mathbf{X}}

\newcommand{\bDeltax}{\mathbf{\Delta x}}

\newcommand{\eq}{=}

\def\g2o{$g^2o$}
\def\se3{\mathrm{SE}(3)}
\def\t2v{\mathrm{t2v}}
\def\v2t{\mathrm{v2t}}
\def\flat{\mathrm{flat}}

\def\secref#1{Sec.~\ref{#1}}
\def\figref#1{Fig.~\ref{#1}}
\def\tabref#1{Tab.~\ref{#1}}
\def\eqref#1{Eq.~(\ref{#1})}
\def\algref#1{Alg.~(\ref{#1})}
\def\lineref#1{(line \ref{#1})}

\newcommand\etal{\emph{et al.}}
\algdef{SE}[DOWHILE]{Do}{doWhile}{\algorithmicdo}[1]{\algorithmicwhile\ #1}%

\def\argmax{\mathop{\rm argmax}}
\def\argmin{\mathop{\rm argmin}}

\newcommand*{\SE}{\ensuremath{\mathrm{SE}(3)}}
\newcommand*{\SO}{\ensuremath{\mathrm{SO}(3)}}
\newcommand{\Real}[1]{\ensuremath{\mathbb{R}^{#1}}}

\begin{document}
\title{HiPE: Hierarchical Initialization for Pose Graphs}
\author{Tiziano Guadagnino\hspace{60pt} Luca Di Giammarino \hspace{60pt} Giorgio Grisetti%
	\thanks{Manuscript received: August, 23, 2021; Revised October, 24, 2021; Accepted October, 24, 2021.}
	\thanks{This paper was recommended for publication by Editor Javier Civera upon evaluation of the Associate Editor and Reviewers' comments.}
	\thanks{All authors are with the Department of Computer, Control, and
		Management Engineering ``Antonio Ruberti", 
		Sapienza University of Rome, Rome,
		Italy. Email: {\tt\small\{guadagnino,digiammarino, grisetti\}@diag.uniroma1.it}}%
	\thanks{Digital Object Identifier (DOI): see top of this page.}
}
\markboth{IEEE Robotics and Automation Letters. Preprint Version. Accepted October, 2021}
{Guadagnino \MakeLowercase{\textit{et al.}}: HiPE} 
	\maketitle
	
	
	\begin{abstract}
		%
		Pose graph optimization is a non-convex optimization problem encountered in many areas of robotics perception. Its convergence to an accurate solution is conditioned by two factors: the non-linearity of the cost function in use and the initial configuration of the pose variables. 
		In this paper, we present HiPE, a novel hierarchical algorithm for pose graph initialization. Our approach exploits a coarse-grained graph that encodes an abstract representation of the problem geometry. We construct this graph by combining maximum likelihood estimates coming from local regions of the input. By leveraging the sparsity of this representation, we can initialize the pose graph in a non-linear fashion, without computational overhead compared to existing methods. The resulting initial guess can effectively bootstrap the fine-grained optimization that is used to obtain the final solution. In addition, we perform an empirical analysis on the impact of different cost functions on the final estimate. Our experimental evaluation shows that the usage of HiPE leads to a more efficient and robust optimization process, comparing favorably with state-of-the-art methods.
	\end{abstract}
	\vspace{-1mm}
	\IEEEpeerreviewmaketitle
	\section{Introduction}
	\label{sec:intro}
	\IEEEPARstart{I}{n} robotics perception, the problem of estimating a collection of poses from relative measurements arises in multiple scenarios. These include, among others,
	\textit{Simultaneous Localization And Mapping} (SLAM)~\cite{cadena2016past}, structure from motion~\cite{tron2016survey}, multi-sensor calibration~\cite{esquivel2007calibration} and bundle adjustment~\cite{agarwal2010bundle}. A pose graph is a graphical representation of the problem, where nodes represent poses, while edges encode spatial constraints between pairs of nodes. As the constraints are typically determined by processing exteroceptive observations, they are affected by uncertainty and they need to be represented in a probabilistic sense. In this context, \textit{Pose Graph Optimization} (PGO) is a non-convex maximum likelihood estimation problem, where we seek for the configuration of nodes which is maximally consistent with the probabilistic constraints. It is traditionally solved under the non-linear least squares framework~\cite{robotics9030051}~\cite{g2o2011kummerle}~\cite{SAM2006dellaert}, which iteratively refines an initial configuration of the variables to obtain the final estimate. Intuitively, a good initialization reduces the computational time of the optimization and the risk of convergence to a sub-optimal configuration.
	
	 The de-facto standard in pose graph initialization is the approach proposed by Martinec
	 and Pajdla~\cite{chordalOriginal2007martinec} and experimentally validated by Carlone \textit{et al.}~\cite{chordalAndInitialization2015carlone}. 
	 
	 \begin{figure}[h]
	 	\centering
	 	
	 	\begin{subfigure}[b]{0.48\columnwidth}
	 		\centering
	 		\includegraphics[width=\linewidth]{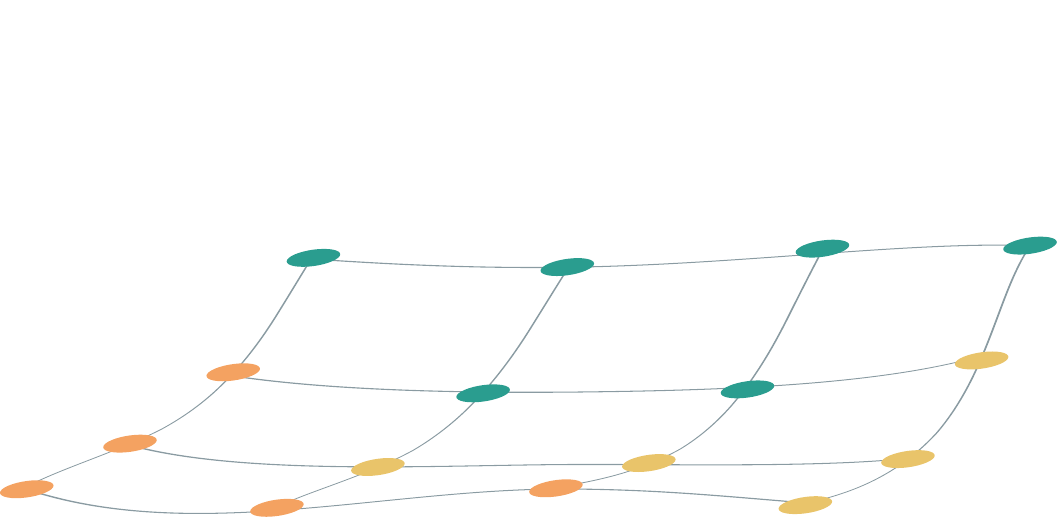}
	 		\caption{}
	 		\label{fig:partitions}
	 	\end{subfigure}
 		\hfill
	 	\begin{subfigure}[b]{0.48\columnwidth}
	 		\centering
	 		\includegraphics[width=\linewidth]{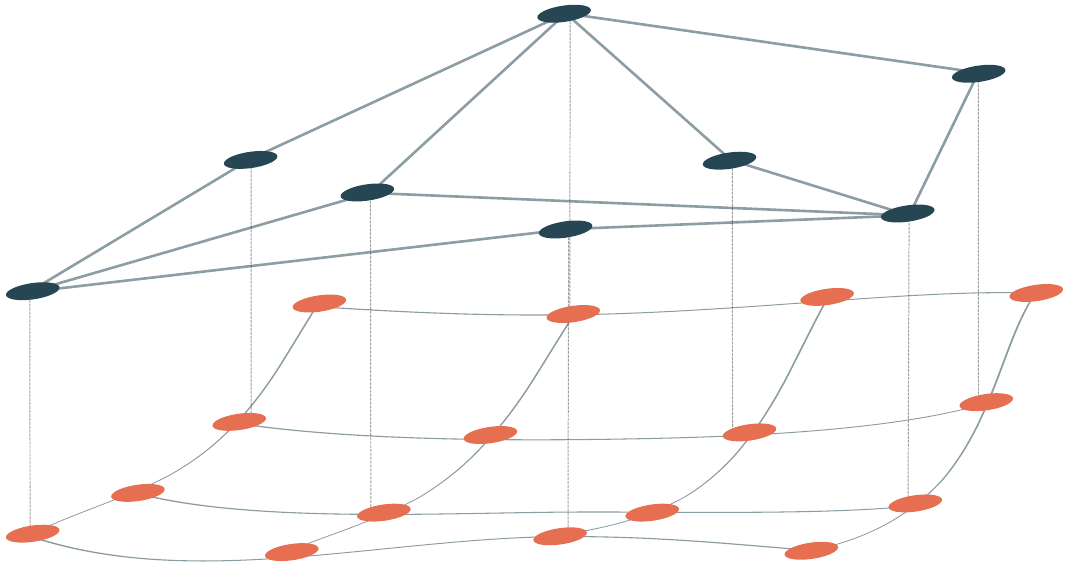}
	 		\caption{}
	 		\label{fig:init}
	 	\end{subfigure}
		\vskip\baselineskip
	 	\begin{subfigure}[b]{0.48\columnwidth}
	 		\centering
	 		\includegraphics[width=\linewidth]{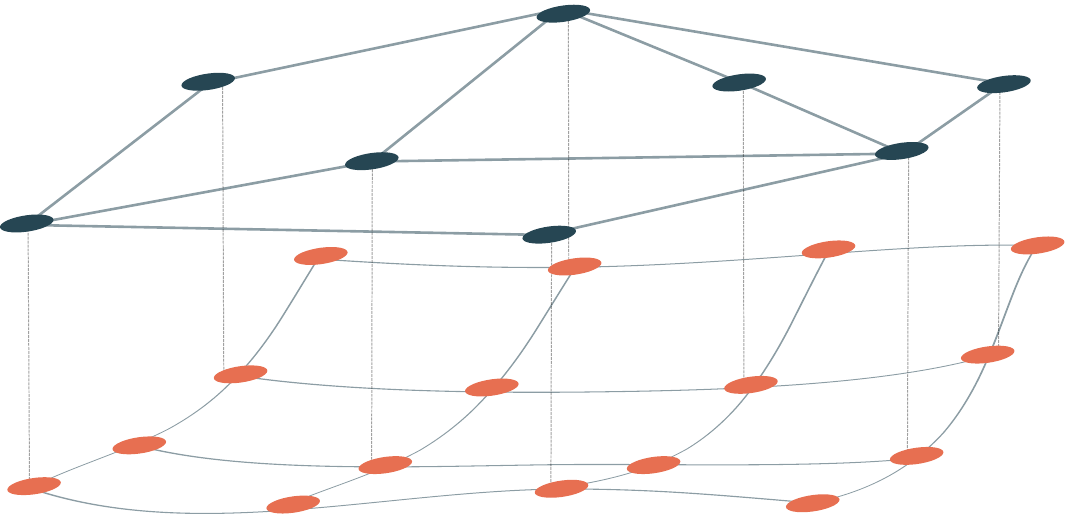}
	 		\caption{}
	 		\label{fig:middle}
	 	\end{subfigure}
	 	\hfill
	 	\begin{subfigure}[b]{0.48\columnwidth}
	 		\centering
	 		\includegraphics[width=\linewidth]{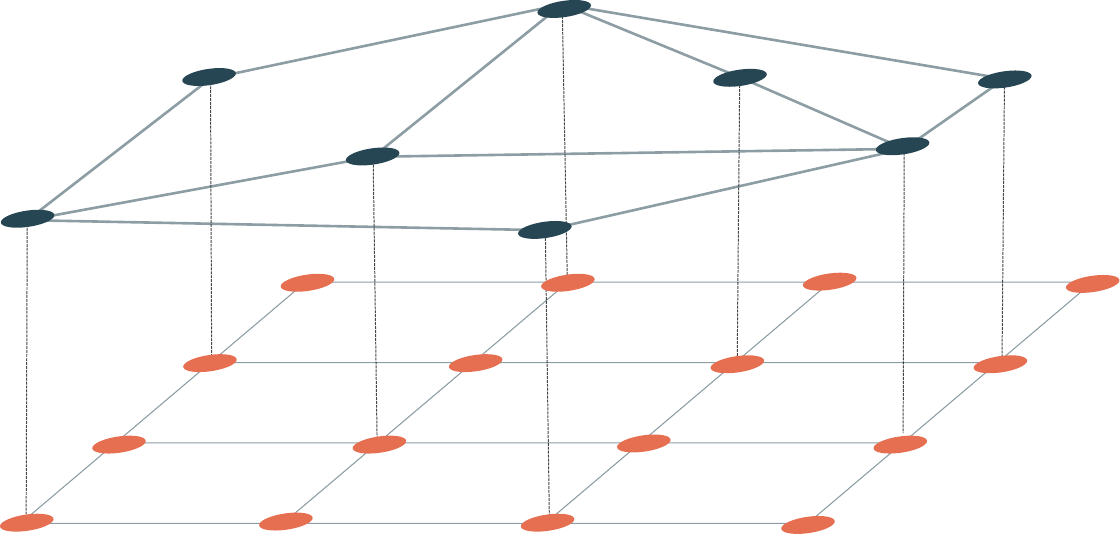}
	 		\caption{}
	 		\label{fig:final}
	 	\end{subfigure}
	 	\caption{HiPE in a nutshell: (\subref{fig:partitions}) the input is partitioned in sub-graphs, (\subref{fig:init}) the coarse-grained graph is constructed using the local estimates of the partitions, (\subref{fig:middle}) the maximum likelihood estimate of the coarse-grained variables is computed, (\subref{fig:final}) the solution is propagated to the remaining variables}
	 	\label{fig:motivational}
	 	\vspace{-1mm}
	 \end{figure}
 This method is based on a linear relaxation of the spatial constraints, which allows to compute an initial configuration of the variables by linear least squares. Even though the relaxation allows an efficient computation, its accuracy rapidly deteriorates with the noise level.
 	In this paper, we present HiPE, a hierarchical initialization approach for pose graphs. Our algorithm is inspired by early works in hierarchical optimization for SLAM~\cite{hogman2010grisetti}~\cite{condensed2012grisetti}~\cite{ni2007tectonic}~\cite{ni2010multi}, which already shows the robustness and efficiency of this type of strategies. Our method partitions the input into a set of sub-graphs and extract local spatial constraints between salient variables. These constraints are then combined in a coarse-grained graph, which represents an abstract topology of the problem. By exploiting a maximum likelihood estimation of this graph, we are able to initialize the pose variables through non-linear optimization. The sparse structure of the coarse-grained representation allows us to efficiently perform the initialization on large-scale problem instances. Further, we exploit rotation estimation approaches~\cite{chordalAndInitialization2015carlone} to bootstrap convergence.\\
	In sum, we make two claims:
	\begin{itemize}
		\item Combining HiPE with a fine-grained optimization delivers performances on par or better than state-of-the-art methods in terms of computational efficiency;
		\item HiPE is highly scalable to large graphs, since it is significantly more accurate and efficient than state-of-the-art methods as the size of the problem grows;
	\end{itemize}
	These claims are supported by our experimental evaluation. In the experiments, we will also show how different assumptions on the distribution of the spatial constraints impact the optimization process. In addition, we release an open-source implementation of HiPE\footnote{ \url{https://github.com/srrg-sapienza/srrg2-hipe}} and the datasets used for the evaluation.
	
	In the following, we will present the approach focusing on the 3D case since closed-form initialization algorithm based on rotation estimation~\cite{carlone2014angular} already exist for 2D pose graphs.

	\section{Related Work}
	\label{sec:related}
	Pose graph optimization was first formulated as a non-linear least squares problem in the seminal paper by Lu~\etal~\cite{lu1997globally}. Even though the \textit{Gauss-Newton} (GN) algorithm was already known from theory, the lack of an efficient numerical implementation prevented its usage in real scenarios.
	
	 The problem was initially tackled using stochastic gradient descent~\cite{olson2006fast}~\cite{grisetti2009nonlinear}, due to its efficiency and large basin of convergence to the global optimum. However, performances were bounded by the slow
	 convergence rate. As more efficient linear algebra libraries became available~\cite{davis2016sparsematrix}, the community rapidly shifted back towards non-linear least squares methods. In the context of SLAM, Dellaert \textit{et al.}~\cite{SAM2006dellaert} was the first to show an efficient approach based on the GN algorithm that exploits sparse matrix decomposition. Moreover, the authors showed the relation between the sparsity pattern of the approximate Hessian and the graph topology. Kaess~\etal~\cite{isam2008kaess} extended these ideas and used a sparse QR factorization to incrementally optimize the graph as new nodes and edges were added. By representing the matrix factorization using a novel data structure called the Bayes Tree, iSAM2~\cite{kaess2012isam2} can further identify the sub-graph which is affected by the newly added measurements. 
	
	As the scale and complexity of the problems continued to grow, the community investigated hierarchical strategies to increase robustness~\cite{condensed2012grisetti}~\cite{ni2010multi}, parallelize the computation~\cite{ni2007out} and incrementally update the graph~\cite{hogman2010grisetti}~\cite{ni2007tectonic}. For large scale problems, solving the linear system underlying the GN algorithm constitutes the major computational bottleneck. To reduce the number of iterations required, researchers developed several initialization strategies with the aim of computing an initial guess as close as possible to the optimum. Konolige~\etal~\cite{konolige2010efficient} introduced an initialization strategy based on a minimum spanning tree, while Hu~\etal~\cite{hu2013towards} proposed an approach that relies on M-estimators and the GN algorithm. In the context of PGO, the attention shifted towards rotation estimation as, with known orientations, the problem can be casted into a linear least squares problem~\cite{khosoussi2016cagate}~\cite{chordalAndInitialization2015carlone}. Sharp~\etal~\cite{sharp2004multiview} proposed a closed-form solution in the single-loop case, while Govindu~\etal~\cite{govindu2001combining} formulated a quaternion relaxation that can be solved through homogenous least squares. Martinec and Pajdla~\cite{chordalOriginal2007martinec} proposed to estimate the rotations through chordal relaxation. As outlined by Carlone~\etal~\cite{chordalAndInitialization2015carlone} the latest approach performs extremely well in practice and is considered the standard for pose graph initialization.
	
	Recently, researchers explored novel formulations of PGO to enhance efficiency and performance guaranties. Rosen~\etal~\cite{sesync2019rosen} developed a specialized solver based on semi-definite relaxation which is guaranteed to recover the optimal estimate of the poses under mild assumptions on the measurements noise. Using a similar formulation, Dellaert~\etal~\cite{Dellaert20eccv-shonan} proposed a rotation averaging algorithm which can be used on large scale rotation-only problems. Bai~\etal~\cite{bai2021cycle} proposed a cycle-based approach for PGO. In this new paradigm, relative poses associated with edges are considered as variables and loop constraints are added to mitigate the over-parameterization of the problem. Then, the resulting constrained least squares is solved using sequential quadratic programming. 
	\section{Pose Graph Optimization}
	Pose graph optimization is the problem of estimating the absolute poses of a set of 3D frames, given a collection of noisy measurements of relative position and orientation between them. This problem is typically represented by an undirected graph $\mathcal{G}\eq\{\mathcal{V},\mathcal{E}\}$, were the nodes $i \in \mathcal{V}$ represent the pose variables and the edges $(i,j)\in\mathcal{E}$ represent measurements between pairs of frames. Traditionally, \textit{Maximum Likelihood Estimation} (MLE) is used to find the configuration of variables which is maximally consistent with the measurements by solving:
	\begin{equation}
		\label{eq:mle}
		\mathcal{X}^* \eq \argmax_{\mathcal{X} \in \SE^N} p(\mathcal{Z}|\mathcal{X})
	\end{equation}
	where $\mathcal{X} = \{\mathbf{X}_i \in \SE \,|\, i \in \mathcal{V}\} $ is the set of variables and $\mathcal{Z} = \{\mathbf{Z}_{ij} \in \SE \,|\, (i,j) \in \mathcal{E} \} $ is the set of relative measurements. 
	 Under the assumption that the measurements are independent and identically distributed, we can write:
	\begin{equation}
		\label{eq:mle_independent}
		\mathcal{X}^* \eq \argmax_{\mathcal{X} \in \SE^N} \prod_{(i,j)\in \mathcal{E}} p(\mathbf{Z}_{ij}|\mathcal{X})
	\end{equation}
	Usually, it is assumed that the distribution $p(\mathbf{Z}_{i,j}|\mathcal{X})$ belongs to the (simplified) exponential family, with form:
	\begin{equation}
		\label{eq:exponential_family}
		p(\mathbf{Z}_{ij}|\mathcal{X}) \sim \mathbf{c}_{ij}\exp(-\mathbf{r}_{ij}(\mathcal{X}))
	\end{equation}
	By plugging \eqref{eq:exponential_family} into \eqref{eq:mle_independent} and taking the negative logarithm we can write:
	\begin{equation}
		\label{eq:least_squares}
		\mathcal{X}^* \eq \argmin_{\mathcal{X} \in \SE^N} \sum_{(i,j)\in\mathcal{E}} \mathbf{r}_{ij}(\mathcal{X})
	\end{equation}
	\eqref{eq:least_squares} is highly non-convex in general, so it has to be solved using iterative unconstrained minimization algorithms. Starting from an initial configuration of the variables, these approaches iteratively refine the estimate by minimizing a linear (or quadratic) approximation of the cost computed using Taylor expansion. Intuitively, the closer the initial configuration is to the optimum the more likely the algorithm can converge to it. At the same time, highly non-linear functions are not well approximated by the Taylor expansion, resulting in poor convergence properties of the algorithms.
	
	\section{Cost functions in Pose Graph Optimization}
	Inspecting \eqref{eq:exponential_family}, we can observe how the properties of the cost function are directly related to the measurements distribution, $p(\mathbf{Z}_{ij}|\mathcal{X})$. In a nutshell, we can vary the convergence properties of our optimization algorithm using different assumptions on the noise affecting the measurements. In the following subsections, we will present the most common choices of this distribution, together with some properties of the corresponding cost function.
	\vspace{-10pt} 
	\subsection{Gaussian distribution over a minimal euclidean parameterization}
	\label{sec:geodesic_sec}
	The most common choice in MLE is to assume Gaussian distributed measurements. In the case of PGO, we need to find a suitable representation of a Gaussian over a manifold. In this sense, we can exploit the smoothness of $\SE$ to locally parameterize a Gaussian distribution around one of its elements. In particular, the mean will be a $\SE$ element, while the covariance matrix will be defined over the Lie Algebra. The resulting local Gaussian distribution around $\mu \in \SE$ can be defined as:
	\begin{equation}
		\label{eq:se3_gaussian}
		\mathcal{N}_{\SE}(\mathbf{x};\small\mathbf{\mu}, \mathbf{\Sigma}) = \frac{\exp(-\frac{1}{2}\:\|Log(\mathbf{x}^{-1}\mu)\|^2_{\mathbf{\Sigma}})}{\sqrt{(2\pi)^6 \det(\mathbf{\Sigma})}}
	\end{equation}
	where $\mu$ is the mean, $\mathbf{\Sigma} \in \mathbb{R}^{6\times6}$ is the covariance matrix, $Log : \SE \rightarrow \mathbb{R}^6$ is the logarithmic mapping, and: 
	\begin{equation}
		\|\mathbf{a}\|^2_{\mathbf{\Sigma}} \eq \mathbf{a}^{T} \mathbf{\Sigma}^{\dagger} \mathbf{a}   
	\end{equation}
	is the squared Mahalonobis distance and $\cdot^{\dagger}$ denotes the Moore–Penrose inverse.\\
	By using \eqref{eq:se3_gaussian}, we can define our distribution of interest as:
	\begin{equation}
		\label{eq:gaussian_se3_prediction}
		p(\mathbf{Z}_{ij}|\mathcal{X}) \propto \exp(-\frac{1}{2}\:\|Log(\mathbf{Z}_{ij}^{-1}\widetilde{\mathbf{Z}}_{ij}(\mathcal{X}))  \|^2_{\mathbf{\Sigma}_{ij}})
	\end{equation}
	
	where $\widetilde{\mathbf{Z}}_{ij} : \SE^N \rightarrow \SE$ determines the mean of the distribution given the current value of the variables. In practice, the mean is conditioned just on the pair of poses involved in the relative measurement, so that:
	\begin{equation}
		\widetilde{\mathbf{Z}}_{ij}(\mathcal{X}) \eq \mathbf{X}_i^{-1}\:\mathbf{X}_j
	\end{equation}
	By inspecting \eqref{eq:gaussian_se3_prediction} we obtain:
	\begin{equation}
		\label{eq:geodesic_residual}
		\mathbf{r}_{i,j}(\mathcal{X}) \eq \frac{1}{2}\:\|Log(\mathbf{Z}_{ij}^{-1}\widetilde{\mathbf{Z}}_{ij}(\mathcal{X}))\|^2_{\mathbf{\Sigma}_{ij}}
	\end{equation}
	which we will refer to as the \textit{Geodesic cost} in the rest of this work. 
	
	While being the most widely used cost function for PGO,  \eqref{eq:geodesic_residual} is highly non-convex due to the usage of the logarithmic mapping. In fact, this function involves multiple non-linear operations to extract a minimal representation of the rotational part of an $\SE$ element.
	\vspace{-5pt}
	\subsection{Gaussian distribution over a non-minimal euclidean parameterization}
	\vspace{1pt}
	\label{sec:chordal_sec}
	Aloise~\etal~\cite{chordal2020aloise} propose a 12-dimensional over-parameterization of the local Gaussian distribution \eqref{eq:se3_gaussian}. In particular, they relax the constraints of $\SE$ on the measurements and interpret them as euclidean objects. Given a transformation matrix:
	\begin{equation}
	\vspace{10pt}
		\mathbf{Z} = \begin{bmatrix}
			\begin{bmatrix} \mathbf{r}_1 & \mathbf{r}_2 & \mathbf{r}_3\end{bmatrix} & \mathbf{t}\\
			\mathbf{0}_{1\times3} & 1
		\end{bmatrix}\in \SE
	\end{equation}
	they define the corresponding euclidean measurement using the $\flat$ operator as:
	\begin{equation}
		\label{eq:flatten}
		\mathbf{z} = \flat(\mathbf{Z})= \begin{pmatrix}
			\mathbf{r}_1\\
			\mathbf{r}_2\\
			\mathbf{r}_3\\
			\mathbf{t}
		\end{pmatrix}\in\mathbb{R}^{12}
	\end{equation}
	Once the measurements are projected in $\mathbb{R}^{12}$, one needs to define a covariance matrix in this new space. A straightforward solution is to perform a first order propagation of the covariance defined over the Lie algebra:
	\begin{equation}
		\label{eq:error_propagation}
		\mathbf{\Sigma}^{[c]} = \mathbf{J} \, \mathbf{\Sigma} \, \mathbf{J}^{T}
	\end{equation}
	where $\mathbf{J} \in \mathbb{R}^{12\times6}$ is the jacobian of \eqref{eq:flatten} with respect to a minimal euclidean parameterization of $\SE$. Notice that the resulting covariance matrix will be rank-deficient by construction.\\
	The over-parametrized Gaussian distribution can be defined as:
	\begin{equation}
		\label{eq:chordal_irvin}
		p(\mathbf{Z}_{ij}|\mathcal{X}) \propto \exp(- \frac{1}{2}\: \|\flat(\widetilde{\mathbf{Z}}_{ij}(\mathcal{X})) - \flat(\mathbf{Z}_{ij})\|^2_{\mathbf{\Sigma}^{[c]}_{ij}})
	\end{equation}
	By simple inspection of \eqref{eq:chordal_irvin}:
	\begin{equation}
		\label{eq:irvin_chordal_residual}
		\mathbf{r}_{ij}(\mathcal{X}) \eq \frac{1}{2}\:\|\flat(\widetilde{\mathbf{Z}}_{ij}(\mathcal{X})) - \flat(\mathbf{Z}_{ij}))  \|^2_{\mathbf{\Sigma}^{[c]}_{ij}}
	\end{equation}
	which will be referred to as the \textit{Chordal cost}.
	
	The intuition behind this approach is that by projecting the measurements into a higher dimensional space, we reduce the non-linearity of the corresponding cost function~\cite{chordal2020aloise}.\\
	\vspace{-10pt}
	\subsection{Combining a Gaussian distribution and an isotropic Langevin distribution}
	\label{sec:se_sync_sec}
	Rosen~\etal~\cite{sesync2019rosen} propose a decoupling of the measurements distribution over $\SE$ as: 
	\begin{align}
	p(\mathbf{R}_{ij}|\mathcal{X}) &\sim \text{Langevin}(\mathbf{R}_{ij} ; \widetilde{\mathbf{R}}_{ij}(\mathcal{X}),\,\kappa_{ij})\\
	p(\mathbf{t}_{ij}|\mathcal{X}) &\sim \mathcal{N}(\mathbf{t}_{ij} ; \widetilde{\mathbf{t}}_{ij}(\mathcal{X}),\tau_{ij}^{-1}\mathbf{I}_{3\times3})
	\end{align}
	where $\text{Langevin}()$ stands for the isotropic Langevin distribution. Implicitly, Rosen~\etal~assume that the measurements have independent components (e.g. the uncertainty of the x-component of the translation is uncorrelated with the uncertainty of the y-component). Our distribution of interest in this case will be:
	\begin{equation}
	\label{eq:se_sync_distro}
	p(\mathbf{Z}_{ij}|\mathcal{X}) \propto \begin{Bmatrix}
	\exp(-\frac{\tau_{ij}}{2}\large \|\mathbf{t}_{ij} - \widetilde{\mathbf{t}}_{ij}(\mathcal{X})\|^2)\\ \times \exp(\kappa_{ij}\text{trace}(\widetilde{\mathbf{R}}_{ij}(\mathcal{X})^{T}\mathbf{R}_{ij}))
	\end{Bmatrix}
	\end{equation}
	Using a simple identity for the $\text{trace}$ function, we can define the cost as:
	\begin{equation}
	\label{eq:se_sync_residual}
	\mathbf{r}_{ij}(\mathcal{X}) \eq \frac{\kappa_{ij}}{2}\|\widetilde{\mathbf{R}}_{ij}(\mathcal{X}) - \mathbf{R}_{ij}\|^2_{_F} + \frac{\tau_{ij}}{2}\|\widetilde{\mathbf{t}}_{ij}(\mathcal{X}) -  \mathbf{t}_{ij}\|^2
	\end{equation}
	To solve the PGO problem Rosen~\etal~form a convex semidefinite relaxation of problem \eqref{eq:least_squares} with cost \eqref{eq:se_sync_residual}. The minimizer of this relaxation provides a \textit{globally optimal} solution, as long as the noise magnitude is below a certain bound. Moreover, the optimality of the solution can be verified a posteriori. However, in real-world scenarios, measurements are rarely affected by isotropic noise, especially in the Langevin sense. As such, \eqref{eq:se_sync_distro} is typically an approximation of the original measurements distribution. In those cases, while the solution found by this approach is optimal in the sense of the cost in~\eqref{eq:se_sync_residual}, it does not coincide with the MLE estimate of the variables given the measurements.  
	
	\section{Chordal Initialization}\label{sec:chordal}
	A typical problem in PGO is to find an initial configuration of the variables, as close as possible to the minimum of \eqref{eq:least_squares}. In this context, the most widely used algorithm is the so-called \textit{Chordal Initialization}~\cite{chordalOriginal2007martinec}~\cite{chordalAndInitialization2015carlone}~\cite{khosoussi2016cagate}. The idea behind this approach is to first estimate the rotation matrices through chordal relaxation and then use this estimate to compute the translation vectors. This rough estimation of the poses is then used as an initial guess for the iterative optimization algorithm.
	
	Chordal relaxation allows to estimate the rotation matrices by solving the unconstrained (linear) problem:
	\begin{equation}
	\label{eq:chordal_relaxation_rotation}
	\min_{\{\mathbf{A}_i\in \mathbb{R}^{3\times3}\}} \sum_{(i,j)\in \mathcal{E}} \| \mathbf{R}^{T}_{ij} \mathbf{A}_i - \mathbf{A}_j\|^2_{_F}
	\end{equation}
	and obtain the matrices ${\mathbf{A}_i}$. These are not rotations in general, so we need to enforce the $\SO$ constraints on each of them. We can do that in closed-form using the singular value decomposition $\mathbf{A}\eq\mathbf{U}\mathbf{S}\mathbf{V}^{T}$ and the formula:
	\begin{equation}
	\mathbf{R}^{*} \eq \mathbf{U}\, \text{diag}([1\, 1\, \det(\mathbf{U}\mathbf{V}^{T})])\, \mathbf{V}^{T}
	\end{equation}
	Given the rotation matrices we can estimate the translation vectors by linear least squares:
	\begin{equation}
	\label{eq:chordal_relaxation_translation}
	\min_{\{\mathbf{t}_i\in \mathbb{R}^3 \}} \sum_{(i,j)\in \mathcal{E}} \| \mathbf{t}_j - \mathbf{t}_i - \mathbf{R}_i^{*}\:\mathbf{t}_{ij}\|^2
	\end{equation}
	While being straightforward, the algorithm performs surprisingly well in practice~\cite{chordalAndInitialization2015carlone}. Furthermore, this represents an efficient solution, as the initialization is constituted by two linear problems. 
	
	\section{Our Approach}
	In this section, we present HiPE, our proposed initialization algorithm for pose graphs. The key idea of this approach is to initialize the input using a coarse-grained graph composed of a subset of variables. This graph, which we call the \textit{skeleton}, must contain enough variables to preserve the spatial structure of the problem while being sparser.

	To construct the \textit{skeleton}, we divide the input into partitions that represent spatially connected regions of the problem. For each of these partitions, a chunk of \textit{skeleton} is assembled using the available measurements within the local graph. As such, each of these chunks encode an abstract representation of the local geometry of the input.

	We then proceed with the initialization in two stages. First, we compute the initial guess for the coarse-grained variables through maximum likelihood estimation. This step merges the local information in the chunks to get a consistent estimate of the \textit{skeleton}. In the second step, we propagate this solution to the remaining variables.
	
	All the steps in our approach use a non-linear least squares minimization algorithm as a back-end. At each iteration, this approach computes the update $\bDeltax$ for the estimate by solving the linear system:
	\begin{equation}
	\label{eq:linear_system}
	\bH\:\bDeltax\eq-\mathbf{g}
	\end{equation}
	where $\bH \in \Real{6N\times6N}$ is the approximate Hessian, and $\mathbf{g}\in\Real{6N}$ is the gradient vector. We can solve \eqref{eq:linear_system} efficiently by exploiting the sparsity of the system matrix $\bH$, as well known in literature \cite{SAM2006dellaert}~\cite{isam2008kaess}~\cite{graphSLAM2010grisetti}~\cite{robotics9030051}. In particular, the sparsity pattern of $\bH$ reflects the connectivity of the pose graph.
	
	
	In the next subsections, we will discuss in detail the different steps outlined above.
	\floatname{algorithm}{Algorithm}
	\renewcommand{\algorithmicrequire}{\textbf{Input:}}
	\renewcommand{\algorithmicensure}{\textbf{Output:}}
	\begin{algorithm}[]
		\caption{Breadth-first Partitioner}
		\label{alg:partitioner}
		\begin{algorithmic}[1]
			\Require{ Graph $\mathcal{G}\eq\{\mathcal{X}, \mathcal{Z}\}$, minimum number of variables $k$, minimum distance on the graph $\gamma$} 
			\Ensure{Graph skeleton $\mathcal{S}\eq\{\mathcal{X}_{\mathcal{S}}, \mathcal{Z}_{\mathcal{S}}\}$}
			\ForAll{$\mathbf{X}_i \in \mathcal{X}$} markUnvisited($\mathbf{X}_i$) \EndFor
			\State $\mathcal{X}_{\mathcal{S}} \eq \emptyset$, $\mathcal{Z}_{\mathcal{S}} \eq \emptyset$ 
			\State Initialize empty queue $\mathcal{Q} \eq \emptyset$
			\State $\mathbf{X} \eq$~MaxDegreeNode($\mathcal{X}$, $\mathcal{Z}$)\label{alg:line:root}
			\Do
			\State $\mathcal{X}_{\mathbf{p}}$, $\mathcal{Z}_{\mathbf{p}}$, $\bar{\mathcal{X}}_{\mathbf{B}} \eq $
			BreadthFirstVisit($\mathcal{G}$, $\mathbf{X}$, $k$, $\gamma$)\label{alg:line:bf}
			\State $\mathbf{X}_{a}\eq$ MaxDegreeNode($\mathcal{X}_{\mathbf{p}}$, $\mathcal{Z}_{\mathbf{p}}$)\label{alg:line:fix}
			\State $\bar{\mathcal{X}}_{\mathbf{p}} \eq \mathcal{X}_{\mathbf{p}}  \setminus \mathbf{X}_a$
			\State Solve $
			\mathcal{X}_{\mathbf{p}}^*, \mathcal{X}_{\mathbf{B}}^* \eq \argmax p(\mathcal{Z}_{\mathbf{p}}|\bar{\mathcal{X}}_{\mathbf{p}}, \bar{\mathcal{X}}_{\mathbf{B}}, \mathbf{X}_a)$\label{alg:line:solve}
			\State $\mathcal{Z}_{\mathbf{B}}\eq$ ComputeVirtualMeasurements($\mathbf{X}_a$, $\mathcal{X}_{\mathbf{B}}^*$)\label{alg:line:vf}
			\State $\mathcal{X}_{\mathbf{B}}\eq \mathcal{X}_{\mathbf{B}}^* \cup \mathbf{X}_a$
			\State $\mathcal{X}_{\mathcal{S}} \eq \mathcal{X}_{\mathcal{S}} \cup \mathcal{X}_{\mathbf{B}}$ 
			\State $\mathcal{Z}_{\mathcal{S}} \eq \mathcal{Z}_{\mathcal{S}} \cup \mathcal{Z}_{\mathbf{B}}$
			\ForAll{$\mathbf{X}_p \in \mathcal{X}_{\mathbf{p}}\cup \mathcal{X}_{\mathbf{B}}$} markVisited($\mathbf{X}_p$) \EndFor
			\ForAll{$\mathbf{X}_b \in \mathcal{X}_{\mathbf{B}}$} addToQueue($\mathcal{Q}$, $\mathbf{X}_b$) \EndFor
			\State $\mathbf{X} \eq$ popFromQueue($\mathcal{Q}$)
			\doWhile{$\mathcal{Q}\neq \emptyset$}
		\end{algorithmic}
	\end{algorithm}
	\vspace{-10pt}
	\subsection{Graph Partitioning}
	
		Our partitioning strategy is presented in \algref{alg:partitioner}. The core of the algorithm is a limited breadth-first visit of the graph~\cite{russel2020modern} that is used to form the partitions~\lineref{alg:line:bf}.
		  In particular, we require a minimum number of visited variables $k$ and a minimum distance traversed on the graph $\gamma$. This is done so that each partition incorporates a sufficient amount of spatial information. 
	  
	  The visit collects variables $\mathcal{X}_{\mathbf{p}}$ and measurements $\mathcal{Z}_{\mathbf{p}}$ that belongs to each partition. Further, it identifies the set of boundary nodes $\mathcal{X}_{\mathbf{B}}$ that will be shared between multiple chunks of the \textit{skeleton}. These boundary variables correspond to the frontier nodes of the breadth-first algorithm plus the visited variables from previous iterations.
	  
	Every time a new partition is formed, we optimize it to get a local estimate of its variables~\lineref{alg:line:solve}. 
	 As we have just relative measurements available, we need to fix the value of a variable during optimization. We select this anchor as the node with maximum degree within the local graph~\lineref{alg:line:fix}.
	 The local estimate is then used to form the chunk of the \textit{skeleton}. In particular, we want to capture the (local) spatial relationships between boundary variables while ignoring all the others. To this end, we add virtual measurements between the anchor and each of the boundary variables, resulting in a sparse approximation of the local graph~\lineref{alg:line:vf}. 
 
 	 We characterize a virtual measurement between boundary variable $\bX_b$ and anchor $\bX_a$ using the relative pose $ \mathbf{Z}_{\mathcal{S}_{ab}} $ between the two frames and a covariance matrix $\mathbf{\Sigma}_{\mathcal{S}_{ab}}$ that quantifies the uncertainty associated with the measurement. In particular, the relative pose can be computed as:
	\begin{equation}
 		\mathbf{Z}_{\mathcal{S}_{ab}} \eq \bX_a^{-1}\:\bX_b
 	\end{equation}

 	while $ \mathbf{\Sigma}_{\mathcal{S}_{ab}} $ corresponds to the marginal covariance of variable $\bX_b$ conditioned on $\bX_a$. We can compute this quantity efficiently using dynamic programming~\cite{kaess2009marginal}.\\
	The overall \textit{skeleton }will be formed by the set of anchors and boundary variables $\mathcal{X}_{\mathcal{S}}$ and corresponding virtual measurements $\mathcal{Z}_{\mathcal{S}}$.
	\subsection{Skeleton Optimization}
	
\begin{figure}[ht]
		\centering
		\begin{subfigure}[b]{\linewidth}
			\centering
			\includegraphics[scale=0.4]{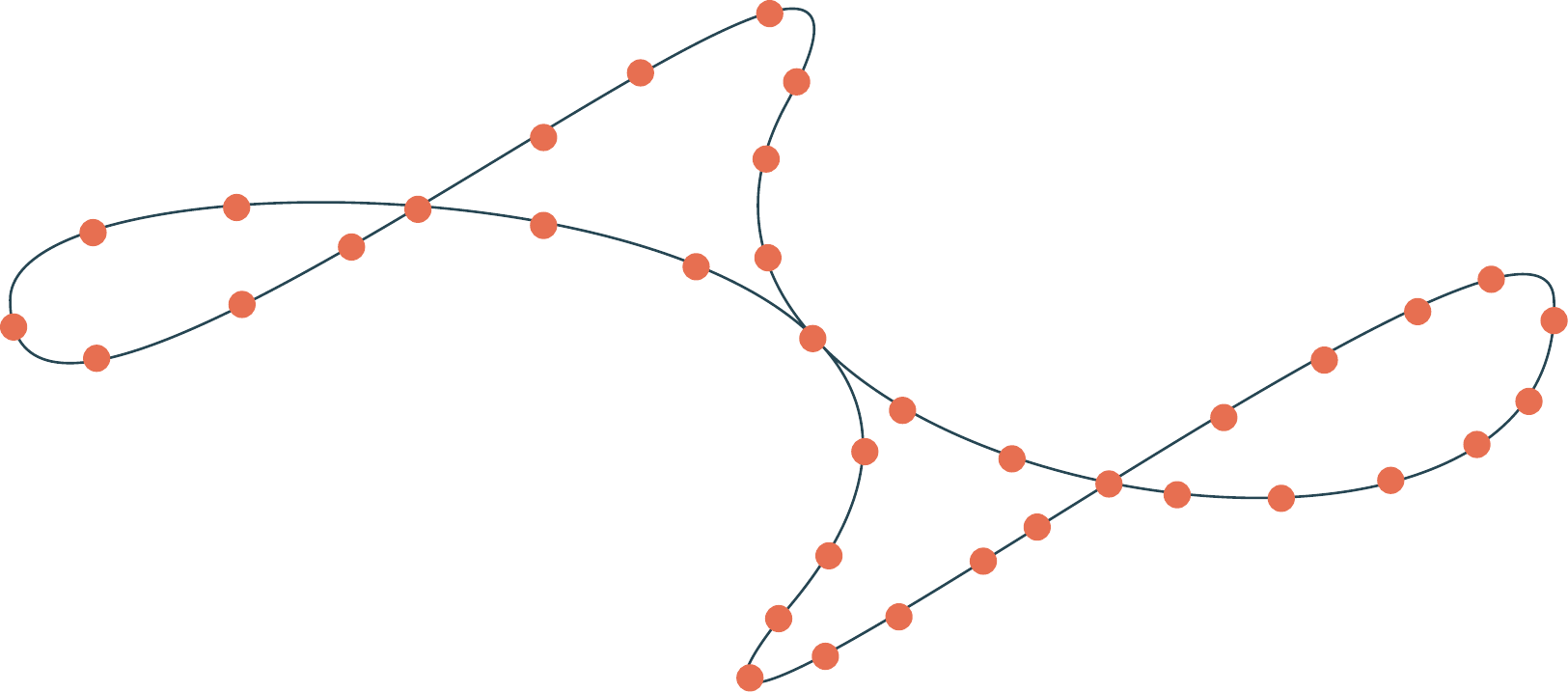}
			\caption{}
			\label{fig:y equals x}
		\end{subfigure}
		\begin{subfigure}[b]{\linewidth}
			\centering
			\includegraphics[scale=0.4]{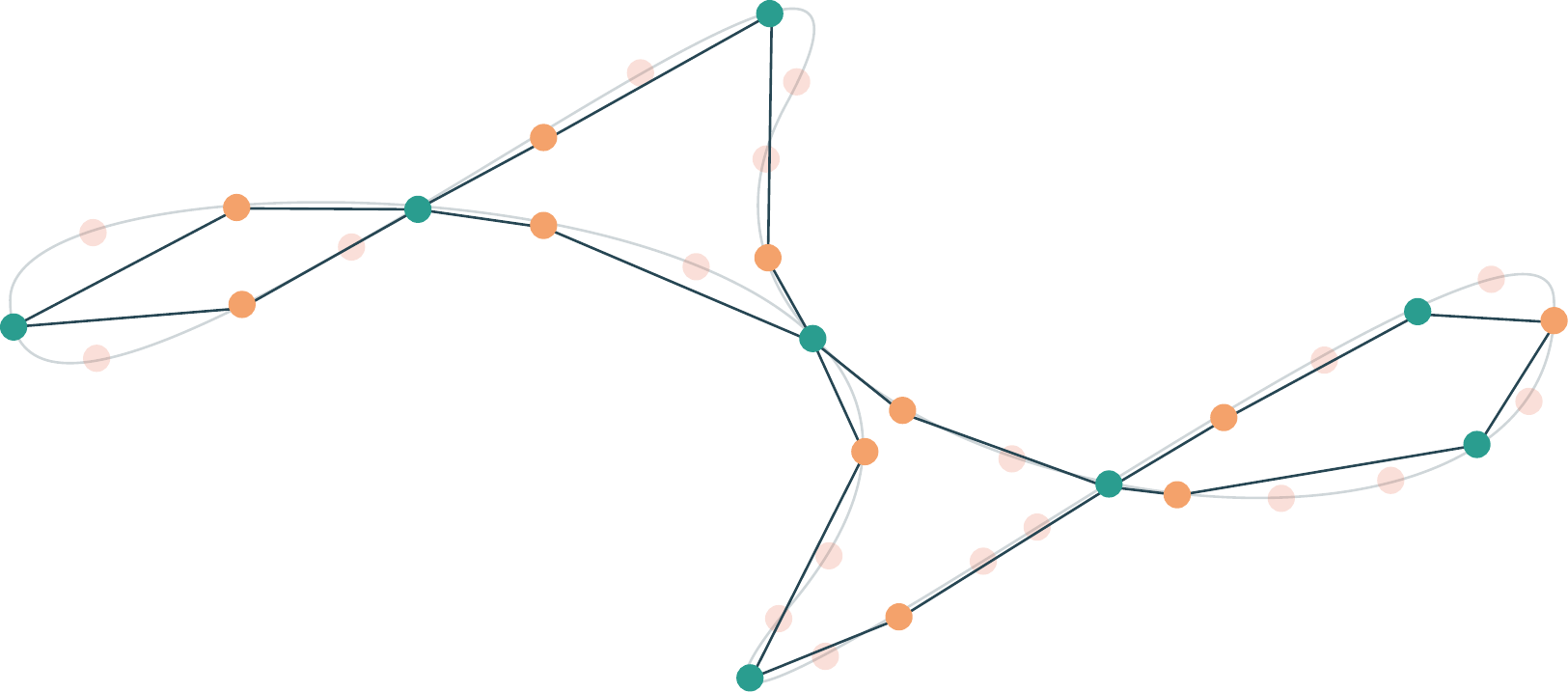}
			\caption{}
			\label{fig:three sin x}
		\end{subfigure}
		\begin{subfigure}[b]{\linewidth}
			\centering
			\includegraphics[scale=0.4]{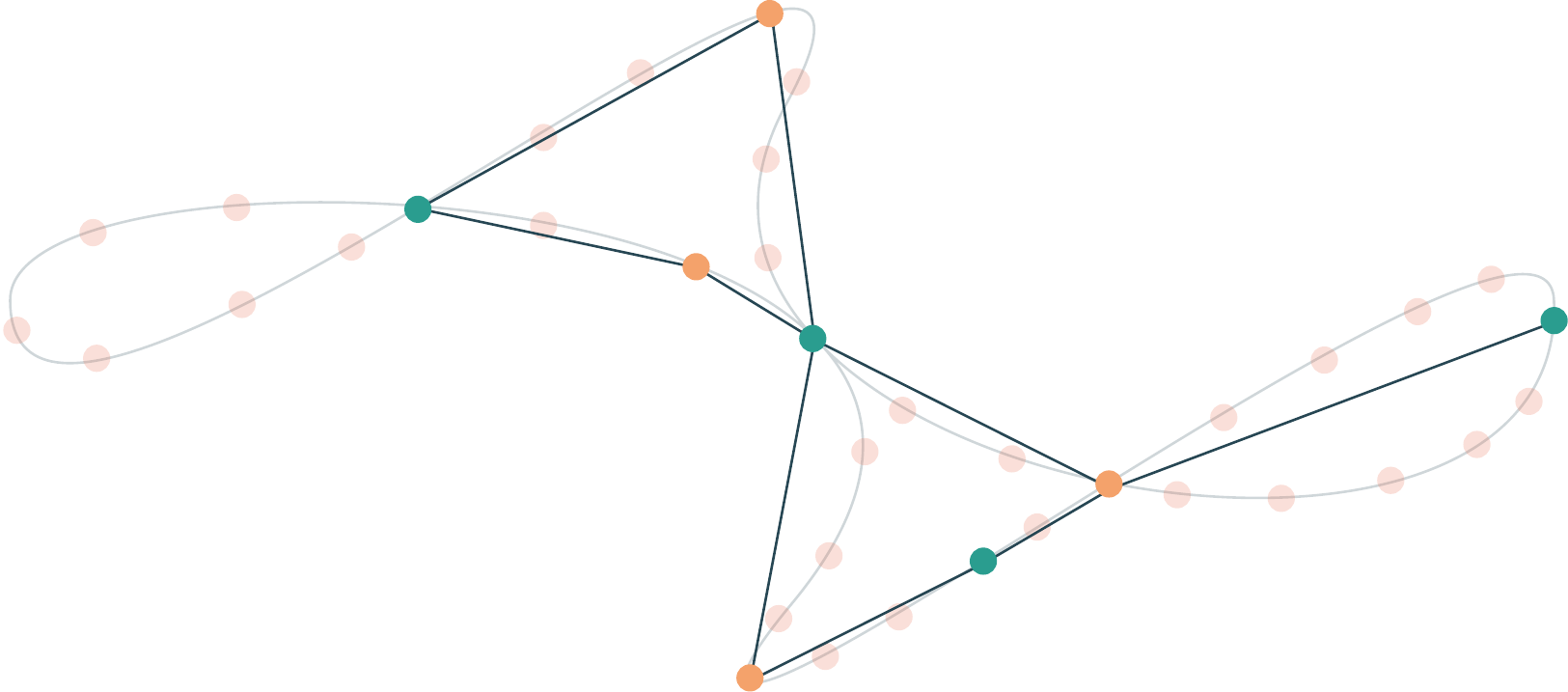}
			\caption{}
			\label{fig:five over x}
		\end{subfigure}
		\caption{(a) The original pose graph,~(b) skeleton obtained with a small partition size,~(c) skeleton obtained with a bigger partition size. In (b) and (c) anchors are in green and boundary variables are in light orange}
		\label{fig:skeleton_partitioning}
		\vspace{-5pt}
	\end{figure}

	The above procedure incrementally creates partitions during the graph visit. As a consequence, boundary variables will have inconsistent estimates due to multiple optimizations in different local graphs. 
	To align the chunks, we compute a maximum likelihood estimate of the \textit{skeleton} variables by solving:
	\begin{equation}
	\label{eq:skeleton}
	\mathcal{X}_{\mathcal{S}}^* \eq \argmax_{\mathcal{X}_{\mathcal{S}}}\,p(\mathcal{Z}_{\mathcal{S}}|\mathcal{X}_{\mathcal{S}})
	\end{equation}
	We initialize $\mathcal{X}_{\mathcal{S}}$ using \textit{Chordal Initialization} to bootstrap convergence. Both steps can be performed efficiently, as the \textit{skeleton} contains just a fraction of variables in the original graph. In particular, the number of variables is controlled by the size of the partitions. As it is intuitive, bigger chunks shift the computation from the \textit{skeleton} to the partitions. At the same time, smaller partitions produce a \textit{skeleton} that is more geometrically consistent with the original graph, improving initialization performances. A visual example is given in \figref{fig:skeleton_partitioning}.   
	\subsection{Optimization of the remaining variables}
	The previous step determines an initial configuration of the \textit{skeleton }variables given the local estimates of the partitions. We would like to propagate this local information to construct an initial guess for the remaining variables, denoted by $\mathcal{X}_{\mathcal{R}}=\mathcal{X}\setminus\mathcal{X}_{\mathcal{S}}$. To do that, we fix the value of the \textit{skeleton} variables and solve the maximum likelihood problem:
	\begin{equation}
	\label{eq:mle_free}
	\mathcal{X}_{\mathcal{R}}^* \eq \argmax_{\mathcal{X}_{\mathcal{R}}}\,p(\mathcal{Z}\,|\,\mathcal{X}_{\mathcal{R}}\text{, }\mathcal{X}_{\mathcal{S}}=\mathcal{X}_{\mathcal{S}}^*)
	\end{equation}
	To boost convergence, we initialize $\mathcal{X}_{\mathcal{R}}$ using \textit{Chordal initialization}.
	Both these operations are extremely efficient as we are fixing the boundary variables of the partitions. As such, we are optimizing independent sub-graphs, which result in a block diagonal structure of the approximate Hessian. 
	\subsection{Fine-grained optimization}
	In the last two steps, we find the MLE of the input graph by considering an approximate topology of anchors and boundary variables. As a result, the solution does not take into account all the measurements at once. We regard this estimate as a robust initialization of the pose graph. To obtain the optimal solution of \eqref{eq:least_squares}, we need to perform a fine-grained optimization using all the available measurements in the input. For this step, any unconstrained non-linear optimization algorithm can be used. 
	\section{Experiments}
	\label{sec:exp}
	The experiments are designed to validate our claims:
	
	\begin{enumerate}
		\item Combining HiPE with a fine-grained optimization delivers performances on par or better than state-of-the-art methods in terms of computational efficiency;
		\item HiPE is highly scalable to large graphs, as it is significantly more accurate and efficient than state-of-the-art methods as the size of the problem grows;
	\end{enumerate}

	Additionally, we will show how the distribution of the measurements influences the convergence properties of the optimization algorithms.
	\subsection{Compared approaches}
	We provide a comparative analysis between HiPE and state-of-the-art approaches. In particular, we select the \textit{Chordal Initialization} method presented in \secref{sec:chordal}, the \textit{Spanning Tree} (SP) approach~\cite{konolige2010efficient} and the \textit{Cauchy Boosting} algorithm (CB)~\cite{hu2013towards}. To evaluate the quality of the initialization, we solve~\eqref{eq:least_squares} starting from the initial guesses computed by each algorithm. We then compare the quality of the final estimates and the total computational time. Further, within the evaluation, we perform an empirical analysis on the impact of different cost functions and optimization algorithms. In particular:
	\begin{itemize}
		\item \textit{Geodesic}: NLS algorithm that minimizes the Geodesic cost~\secref{sec:geodesic_sec};
		\item \textit{Chordal}:  NLS algorithm that minimizes the Chordal cost~\secref{sec:chordal_sec};
		\item \textit{SE-Sync}: The approach described in~\cite{sesync2019rosen}. This method uses a truncated Newton algorithm to minimize the cost presented in \secref{sec:se_sync_sec};
	\end{itemize}
	The evaluation of the initialization strategies is performed separately for each of the above cases. The NLS algorithm and all the initialization algorithms are implemented in C++ using the \textit{srrg2\_solver} framework~\cite{robotics9030051}. For the NLS optimization, we use the Powell-Dogleg algorithm~\cite{powell1970new} and clamp the maximum iterations to $10$. For the SE-Sync algorithm, we leave the default configuration provided with the software package.
	
	 A cost-based termination criterion is used for all the presented algorithms. The SE-Sync method has a further termination criteria based on the global optimality of the solution. We will include this step in the timing comparison, as the algorithm will proceed with the optimization if the solution is a saddle point.
	\subsection{Benchmarks and metrics}
We perform the evaluations on both publicly available 
dataset and own ones, in particular we select:

\begin{itemize}
	\item \textit{grid3D} and \textit{torus3D} which are publicly available synthetic datasets;
	\item \textit{parking-garage}, a publicly available real-world dataset;
	\item \textit{rim} and \textit{cubicle} which are real-world datasets. In particular, we select the version with isotropic noise model;
	\item \textit{sphere5000} that we generated by simulating a vehicle moving on the surface of a sphere. To make the dataset challenging we injected an extremely high Gaussian noise on the rotation ($\sigma_{\theta}\eq 0.6 [\text{rad}]$);
\end{itemize}
We released these datasets to make the experiments repeatable by the community.
	 
	To evaluate the quality of the estimate, we use the normalized $\chi^2$ metric, defined as:
	\begin{equation}
		\chi^2(\mathcal{X}) \eq \frac{1}{6\,(m-n)}\sum_{(i,j)\in\mathcal{E}} \mathbf{r}_{ij}(\mathcal{X})
	\end{equation}
	where $m$ is the number of measurements and $n$ is the number of variables. In addition, we report the number of iterations of the fine-grained optimizer and the cumulative time (initialization + optimization).
	
	All the experiments are performed on a Dell XPS 15 laptop with an Intel Core i7-10750H and 16 GB of RAM running Ubuntu 20.04.
			
\begin{figure}[h]
	\centering
	\includegraphics[width=0.9\linewidth]{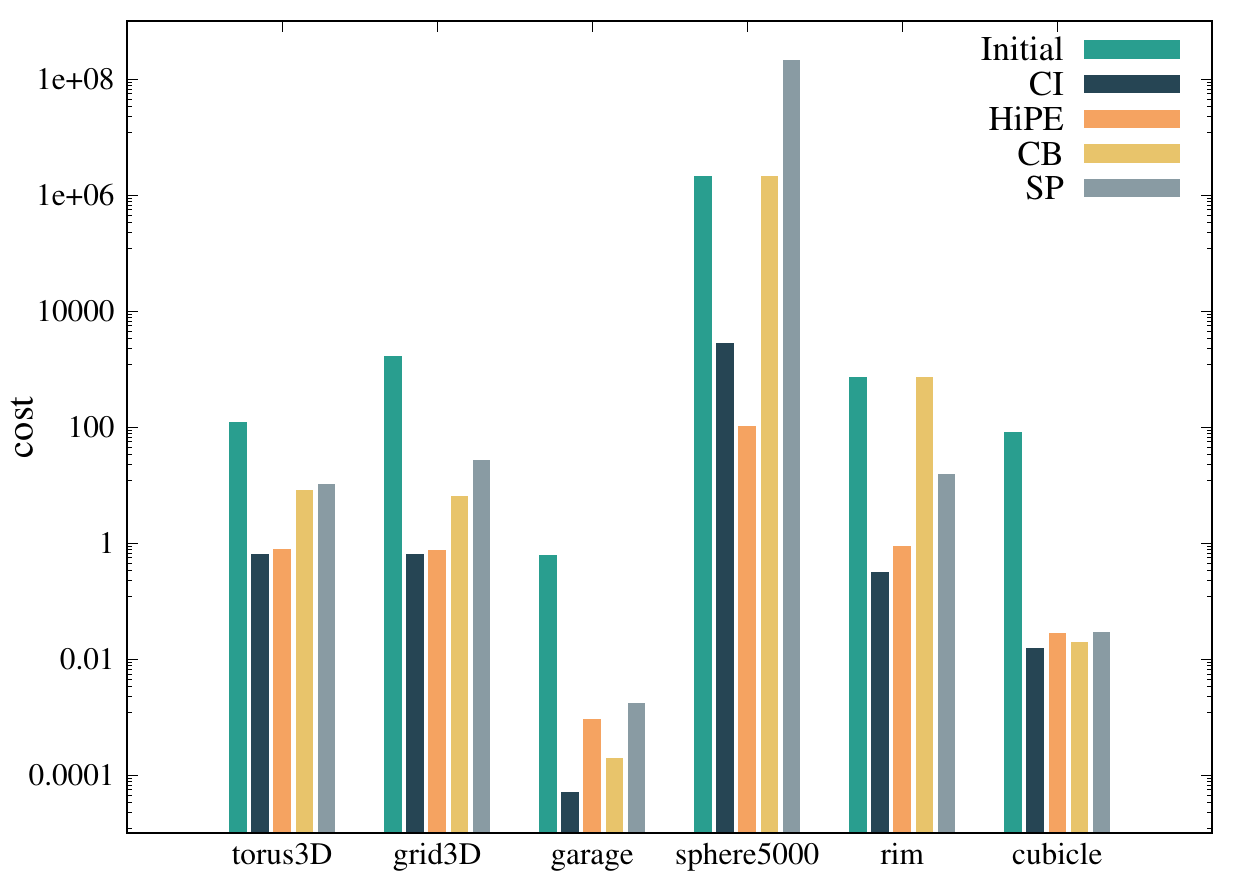}
	\caption{Normalized $\chi^2$ after initialization using the Geodesic cost. \textit{Initial} stands for the odometry configuration given with the datasets}
	\label{fig:benchmarks}
\end{figure}
	\subsection{Performances}
	\label{sec:performances}
	The first set of experiments is designed to show the performance of our approach. 
	The experimental evaluation is presented in \tabref{tab:results}, while in~\figref{fig:benchmarks} we report the normalized $\chi^2$ value after initialization. The most impressive result is obtained in the \textit{sphere5000} dataset, where the usage of HiPE leads to a substantial improvement in estimation accuracy and computational time. In this particular benchmark, we can see how, by exploiting a coarse-grained representation, our proposed initialization strategy is effective in high noise regimes, no matter the cost function used in the fine-grained optimization.
	
	Another interesting result is obtained in \textit{grid3D}, which is the graph with the highest density of edges. In this benchmark, the proposed initialization strategy significantly improves computational efficiency. The higher density of measurements in this dataset makes the CI and CB methods computationally demanding, as the corresponding linear systems are dense. Instead, due to the sparse structure of the \textit{skeleton}, HiPE can exploit non-linear optimization efficiently without compromising initialization quality.

	In the \textit{garage} dataset, the other methods have a slight advantage over our approach in terms of computation time. However, looking at the number of iterations required by the optimizer, we can see that the results are overall comparable. As this benchmark has the smallest number of variables and measurements, the construction and optimization of the skeleton are slightly more demanding in terms of time compared to a graph visit (SP) and optimization-based strategies (CB and CI).

	For what concern \textit{rim} and \textit{cubicle}, results are overall comparable. In these benchmarks we can see how, depending on the cost function in use, the compared initialization strategies lead to different performances in terms of computational time.

\begin{table*}
	\centering
	\vspace{10pt}
	\begin{tabular}{|cc|cccc|cccc|cccc|} 
		\hline
		\multicolumn{2}{|c}{} & \multicolumn{4}{|c}{Geodesic} & \multicolumn{4}{|c}{Chordal} & \multicolumn{4}{|c|}{SeSync} \\
		& & SP & CB & CI & HiPE & SP & CB & CI & HiPE & SP & CB & CI & HiPE \\
		\hline
		\begin{tabular}[c]{@{}c@{}}torus3D\\ n=5000\\ m=9049\end{tabular} & \begin{tabular}[c]{@{}c@{}}Cost\\ Iter\\ Time\end{tabular} &
		\begin{tabular}[c]{@{}c@{}}0.60\\ 7\\ 4.62\end{tabular}&
		\begin{tabular}[c]{@{}c@{}}1.00\\ 10\\ 12.98\end{tabular}&
		\begin{tabular}[c]{@{}c@{}}0.60\\ 3\\ 3.14\end{tabular}           & \begin{tabular}[c]{@{}c@{}}0.60\\ 3\\\textbf{ 2.84 }\end{tabular} & 
		\begin{tabular}[c]{@{}c@{}}0.07 \\ 6 \\ 3.66 \end{tabular} &
			\begin{tabular}[c]{@{}c@{}}0.16 \\ 10 \\ 13.15\end{tabular} &
			\begin{tabular}[c]{@{}c@{}}0.07 \\ 3 \\ 3.06\end{tabular} &
			\begin{tabular}[c]{@{}c@{}}0.07 \\ 3 \\ \textbf{2.92}\end{tabular} &
		\begin{tabular}[c]{@{}c@{}}1.00\\ 13\\ \textbf{1.23}\end{tabular}&
		\begin{tabular}[c]{@{}c@{}}1.00\\ 58\\ 14.76\end{tabular}&
		\begin{tabular}[c]{@{}c@{}}1.00\\\textbf{ 6}\\ 2.10\end{tabular}           & \begin{tabular}[c]{@{}c@{}}1.00\\ 9\\1.93 \end{tabular}              \\ 
		\hline
		\begin{tabular}[c]{@{}c@{}}grid3D\\ n=8000\\ m=22237\end{tabular}     & \begin{tabular}[c]{@{}c@{}}Cost\\ Iter\\ Time\end{tabular} & 
		\begin{tabular}[c]{@{}c@{}}0.58\\ 7\\ 108.96\end{tabular}&
		\begin{tabular}[c]{@{}c@{}}0.58\\ 4\\ 220.32\end{tabular}&
		\begin{tabular}[c]{@{}c@{}}0.58\\ 3\\ 92.15\end{tabular}          & \begin{tabular}[c]{@{}c@{}}0.58\\ 3\\\textbf{ 51.54 }\end{tabular}                  & 
		\begin{tabular}[c]{@{}c@{}}0.06 \\ 6 \\ 96.67\end{tabular} &
		\begin{tabular}[c]{@{}c@{}}0.06 \\ 5 \\ 229.42\end{tabular} &
		\begin{tabular}[c]{@{}c@{}}0.06 \\ 3 \\ 90.41\end{tabular} &
		\begin{tabular}[c]{@{}c@{}}0.06 \\ 3 \\ \textbf{51.65}\end{tabular} &  
		\begin{tabular}[c]{@{}c@{}}0.99\\ 15\\ 9.66\end{tabular}&
		\begin{tabular}[c]{@{}c@{}}0.99\\ 12\\ 165.94\end{tabular}&
		\begin{tabular}[c]{@{}c@{}}0.99\\\textbf{ 8}\\ 50.23\end{tabular}          & \begin{tabular}[c]{@{}c@{}}0.99\\ 9\\\textbf{9.47}\end{tabular}            \\ 
		\hline
		\begin{tabular}[c]{@{}c@{}}garage\\ n=1661\\ m=6276\end{tabular}      & \begin{tabular}[c]{@{}c@{}}Cost\\ Iter\\ Time\end{tabular} &
		\begin{tabular}[c]{@{}c@{}}4e-5\\ 8\\ 0.35\end{tabular}&
		\begin{tabular}[c]{@{}c@{}}4e-5\\ 7\\ 0.37\end{tabular}&
		\begin{tabular}[c]{@{}c@{}}4e-5\\ 3\\0.29 \end{tabular} & \begin{tabular}[c]{@{}c@{}}4e-5\\ 3\\ 0.49\end{tabular}                             & 
		\begin{tabular}[c]{@{}c@{}}4e-5 \\ 1 \\ \textbf{0.09}\end{tabular} &
		\begin{tabular}[c]{@{}c@{}}4e-5 \\ 1 \\ 0.17\end{tabular} &
		\begin{tabular}[c]{@{}c@{}}4e-5 \\ 1 \\ 0.21\end{tabular} &
		\begin{tabular}[c]{@{}c@{}}4e-5 \\ 1 \\ 0.63\end{tabular} & 
		                      
		\begin{tabular}[c]{@{}c@{}}4e-5\\ 9\\ 3.95\end{tabular}&
		\begin{tabular}[c]{@{}c@{}}4e-5\\ 7\\ 4.23\end{tabular}&
		\begin{tabular}[c]{@{}c@{}}4e-5\\\textbf{ 5}\\\textbf{ 2.24 }\end{tabular} & \begin{tabular}[c]{@{}c@{}}4e-5\\ 8\\ 4.76\end{tabular}                        \\ 
		\hline
		\begin{tabular}[c]{@{}c@{}}sphere5000\\ n=5000\\ m=19800\end{tabular} & \begin{tabular}[c]{@{}c@{}}Cost\\ Iter\\ Time\end{tabular} &
		\begin{tabular}[c]{@{}c@{}}775.47\\ 10\\ 12.29\end{tabular}&
		\begin{tabular}[c]{@{}c@{}}3165.34\\ 10\\ 15.22\end{tabular}&
		\begin{tabular}[c]{@{}c@{}}435.32\\ 10\\ 16.17\end{tabular}       & \begin{tabular}[c]{@{}c@{}}\textbf{1.33}\\\textbf{ 5}\\\textbf{9.11}\end{tabular} & 
		\begin{tabular}[c]{@{}c@{}}70.61 \\ 10 \\ 12.40\end{tabular} &
		\begin{tabular}[c]{@{}c@{}}35711.60 \\ 1 \\ 5.53\end{tabular} &
		\begin{tabular}[c]{@{}c@{}}29.14 \\ 10 \\ \textbf{10.90}\end{tabular} &
		\begin{tabular}[c]{@{}c@{}}\textbf{0.46} \\ 10 \\ 11.99\end{tabular} & 
		
		\begin{tabular}[c]{@{}c@{}}4.25\\ 344\\ 478.79\end{tabular}&
		\begin{tabular}[c]{@{}c@{}}4.25\\ 259\\ 516.09\end{tabular}&
		\begin{tabular}[c]{@{}c@{}}4.25\\ 205\\ 372.02\end{tabular}                & \begin{tabular}[c]{@{}c@{}}4.25\\\textbf{ 17}\\\textbf{ 96.80 }\end{tabular}  \\ 
		\hline
		\begin{tabular}[c]{@{}c@{}}rim\\ n=10195\\ m=29744\end{tabular}       & \begin{tabular}[c]{@{}c@{}}Cost\\ Iter\\ Time\end{tabular} & 
		\begin{tabular}[c]{@{}c@{}}0.20\\ 8\\ 8.50\end{tabular}&
		\begin{tabular}[c]{@{}c@{}}0.21\\ 10\\ 11.73\end{tabular}&
		\begin{tabular}[c]{@{}c@{}}0.20\\ 5\\ 7.93\end{tabular}           & \begin{tabular}[c]{@{}c@{}}0.20\\ 5\\\textbf{ 7.42 }\end{tabular}                   & 
\begin{tabular}[c]{@{}c@{}}0.01 \\ 7 \\ \textbf{7.29}\end{tabular} &
\begin{tabular}[c]{@{}c@{}}0.03 \\ 10 \\ 11.55\end{tabular} &
\begin{tabular}[c]{@{}c@{}}0.01 \\ 6 \\ 8.89\end{tabular} &
\begin{tabular}[c]{@{}c@{}}0.01 \\ \textbf{5} \\ 8.02\end{tabular} & 
 
		\begin{tabular}[c]{@{}c@{}}0.31\\ 29\\ 49.88\end{tabular}&
		\begin{tabular}[c]{@{}c@{}}0.31\\ 135\\ 153.18\end{tabular}&
		\begin{tabular}[c]{@{}c@{}}0.31\\\textbf{ 10}\\ 38.23\end{tabular}         & \begin{tabular}[c]{@{}c@{}}0.31\\ 13\\\textbf{ 34.61 }\end{tabular}            \\ 
		\hline
		\begin{tabular}[c]{@{}c@{}}cubicle\\ n=5750\\ m=16870\end{tabular}    & \begin{tabular}[c]{@{}c@{}}Cost\\ Iter\\ Time\end{tabular} &
		\begin{tabular}[c]{@{}c@{}}0.01\\ 7\\ 2.94\end{tabular}&
		\begin{tabular}[c]{@{}c@{}}0.01\\ 4\\ 5.76\end{tabular}&
		\begin{tabular}[c]{@{}c@{}}0.01\\ 3\\ 2.52\end{tabular}           & \begin{tabular}[c]{@{}c@{}}0.01\\ 3\\ 2.55\end{tabular}                             & 
		\begin{tabular}[c]{@{}c@{}}0.01 \\ 7 \\ 2.81\end{tabular} &
		\begin{tabular}[c]{@{}c@{}}0.01 \\ 5 \\ 6.14\end{tabular} &
		\begin{tabular}[c]{@{}c@{}}0.01 \\ \textbf{4} \\ \textbf{2.72}\end{tabular} &
		\begin{tabular}[c]{@{}c@{}}0.01 \\ 5 \\ 3.02\end{tabular} & 
		
		\begin{tabular}[c]{@{}c@{}}0.02\\ 19\\ 9.67\end{tabular}&
		\begin{tabular}[c]{@{}c@{}}0.02\\ 7\\ 9.99\end{tabular}&
		\begin{tabular}[c]{@{}c@{}}0.02\\\textbf{ 6}\\ \textbf{6.42} \end{tabular} & \begin{tabular}[c]{@{}c@{}}0.02\\ 9\\ 6.81\end{tabular}                        \\
		\hline
		
	\end{tabular}
	
\caption{Comparative analysis on the selected benchmarks. With reference to \algref{alg:partitioner} we use $k=100$, $\gamma=50$.}
\label{tab:results}
\end{table*}
	\subsection{Cost functions analysis}
	In the previous experiments, we can observe how different assumptions on the measurement distribution impact the convergence behaviour of the optimization algorithms. In particular, each combination of initialization strategy and cost function lead to a substantial change in computational performances. For example, SP significantly improves its effectiveness when combined with the \textit{Chordal cost}, as we can see in the \textit{rim} and \textit{garage} benchmarks. Even though this initialization strategy typically provides a poor initial guess in datasets with multiple loops, combining its efficiency with the smoothness of the \textit{Chordal cost} delivers good performances in terms of computational time. Instead, HiPE achieves the best performances when combined with the \textit{Geodesic cost}, as the proposed strategy better handles non-linearities compared to other state-of-the-art methods.

	For what concern SE-Sync, it is worth noticing the distinction between isotropic Gaussian and Langevin~\cite[Appendix A]{sesync2019rosen}. Moreover, approximating a Gaussian distribution with a Langevin one typically leads to an overconfident estimation of the uncertainty and thus, a relaxed version of the corresponding optimization problem. As all the considered benchmarks have a Gaussian measurement distribution, the variables configuration found by SE-Sync differs from the MLE computed using the \textit{Geodesic} cost. To show that, we report in \tabref{tab:ate_sesync} the \textit{Absolute Trajectory Error} between the two estimates for all the selected datasets. As we can see, the two solutions are significantly different. Nevertheless, this approach has impressive robustness to noise independently from the initialization strategy in use, as we can see in \textit{sphere5000}. Moreover, it is the only method that can verify the optimality of the solution a posteriori\footnote{For the benchmarks, SE-Sync verify the optimality of the returned solution in all the cases except \textit{sphere5000}}. 
	
	\begin{table}[h]
		\centering
		\begin{tabular}{|c|c|c|}
			\hline
			& ATE [rad] & ATE [m] \\
			\hline
			torus3D & $0.0303$ & $0.2036$ \\
			\hline
			grid3D & $0.0255$ & $0.4083$ \\
			\hline
			garage & $0.0088$ & $1.6175$ \\
			\hline
			sphere5000 & $0.0226$ & $3.3795$ \\
			\hline
			rim & $0.0414$ & $0.3781$ \\
			\hline
			cubicle & $0.0104$ & $0.1301$ \\
			\hline
		\end{tabular}
		\caption{\textit{Absolute Trajectory Error}~(ATE) of the SE-Sync estimate with respect to the MLE estimate}
		\label{tab:ate_sesync}
	\end{table}
	
	\begin{figure}[h]
		\centering
		\includegraphics[width=0.85\linewidth]{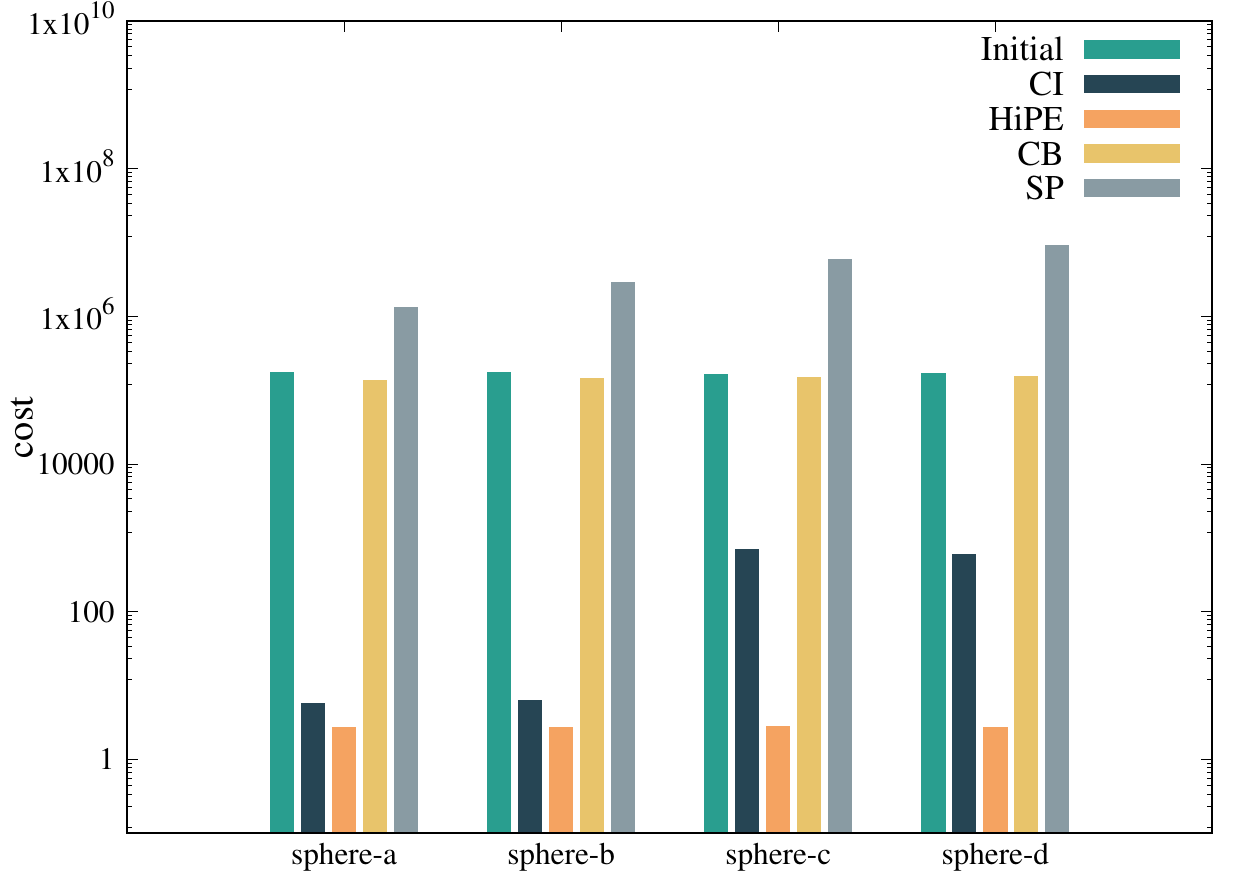}
		\caption{Normalized $\chi^2$ after initialization using the Geodesic cost. \textit{Initial} corresponds to the odometry configuration.}
		\label{fig:chi_init}
	\vspace{-10pt}
	\end{figure}
	
	\subsection{Scalability}
	\label{subsec:scalability}
	In the second experiment, we aim to support our claim that HiPE is highly scalable to large graphs. To do that, we generate four sphere-like datasets with an increasing number of variables and measurements using \textit{g2o}~\cite{g2o2011kummerle}.
	\begin{itemize}
		\item \textit{sphere-a} : 10000 variables 39799 measurements;
		\item \textit{sphere-b} : 20000 variables 79799 measurements;
		\item \textit{sphere-c} : 40000 variables 159799 measurements;
		\item \textit{sphere-d} : 80000 variables 309799 measurements;
	\end{itemize}
	In all of these dataset we inject Gaussian noise on both rotation ($\sigma_{\theta}\eq 0.03$[rad]) and translation ($\sigma_t\eq0.01$[m]).
	A comparative analysis with state-of-the-art methods is reported in~\figref{fig:chi_init}  and~\tabref{tab:scalability_timings}. The analysis shows that HiPE is more efficient and accurate than state-of-the-art methods as the size of the graph grows. This is due to the capability of the proposed method to better handle the additional non-linearities resulting from the increased number of spatial constraints in the graphs.
	\begin{table}[h]
		\centering
		\vspace{10pt}
	\begin{tabular}{|cc|cccc|} 

		\hline
		& & SP & CB & CI & HiPE  \\
		\hline
	\begin{tabular}[c]{@{}c@{}}sphere-a\\ \end{tabular} &
	\begin{tabular}[c]{@{}c@{}}Cost\\ Iter\\ Time\end{tabular} &
	\begin{tabular}[c]{@{}c@{}}1.01 \\ 10 \\ 22.97\end{tabular} &
	\begin{tabular}[c]{@{}c@{}}1.08 \\ 10 \\ 59.23\end{tabular} &
	\begin{tabular}[c]{@{}c@{}}1.01 \\ 3 \\ 11.09\end{tabular} &
	\begin{tabular}[c]{@{}c@{}}1.01 \\ \textbf{2} \\ \textbf{9.09}\end{tabular} \\ 
	\hline
	\begin{tabular}[c]{@{}c@{}}sphere-b\\ \end{tabular} &
	\begin{tabular}[c]{@{}c@{}}Cost\\ Iter\\ Time\end{tabular} &
	\begin{tabular}[c]{@{}c@{}}1.01 \\ 10 \\ 63.84\end{tabular} &
	\begin{tabular}[c]{@{}c@{}}1.02 \\ 10 \\ 157.48\end{tabular} &
	\begin{tabular}[c]{@{}c@{}}1.01 \\ 4 \\ 28.52\end{tabular} &
	\begin{tabular}[c]{@{}c@{}}1.01 \\ \textbf{2} \\ \textbf{19.03}\end{tabular} \\ 
	\hline
	\begin{tabular}[c]{@{}c@{}}sphere-c\\ \end{tabular} &
	\begin{tabular}[c]{@{}c@{}}Cost\\ Iter\\ Time\end{tabular} &
	\begin{tabular}[c]{@{}c@{}}23.70 \\ 10 \\ 107.35\end{tabular} &
	\begin{tabular}[c]{@{}c@{}}1.32 \\ 10 \\ 253.03\end{tabular} &
	\begin{tabular}[c]{@{}c@{}}112.23 \\ 10 \\ 104.02\end{tabular} &
	\begin{tabular}[c]{@{}c@{}}\textbf{1.01} \\ \textbf{2} \\ \textbf{38.56}\end{tabular} \\ 
	\hline
	\begin{tabular}[c]{@{}c@{}}sphere-d\\ \end{tabular} &
	\begin{tabular}[c]{@{}c@{}}Cost\\ Iter\\ Time\end{tabular} &
	\begin{tabular}[c]{@{}c@{}}8944.29 \\ 10 \\ 218.43\end{tabular} &
	\begin{tabular}[c]{@{}c@{}}2973.68 \\ 10 \\ 492.15\end{tabular} &
	\begin{tabular}[c]{@{}c@{}}262.87 \\ 10 \\ 215.28\end{tabular} &
	\begin{tabular}[c]{@{}c@{}}\textbf{1.01} \\ \textbf{3} \\ \textbf{98.39}\end{tabular}  \\
	\hline
	\end{tabular}
		\caption{Comparative analysis for the scalability experiments using the Geodesic cost. With reference to \algref{alg:partitioner}, we use $k=100$, $\gamma=50$. }
		\label{tab:scalability_timings}
		\vspace{-10pt}
	\end{table}

	\section{Conclusion}
	\label{sec:conclusion}

	In this paper, we presented HiPE: a hierarchical approach to pose graph initialization. Our method exploits a coarse-grained graph that encodes a high-level representation of the graph geometry to robustly initialize the variables.  
	We implemented and evaluated our method on different datasets and provided comparisons with state-of-the-art methods. Experiments show that HiPE leads to a more efficient and robust optimization process, comparing favorably with state-of-the-art methods. In addition, we show how different choices of the measurements distribution impact the convergence of the optimization. Finally, we publicly release our own datasets, as well as an open-source implementation of the approach.

	\bibliographystyle{IEEEtran}
	\bibliography{IEEEabrv,2021_paper_guadagnino_hierarchical}

\begin{thebibliography}{10}
\providecommand{\url}[1]{#1}
\csname url@rmstyle\endcsname
\providecommand{\newblock}{\relax}
\providecommand{\bibinfo}[2]{#2}
\providecommand\BIBentrySTDinterwordspacing{\spaceskip=0pt\relax}
\providecommand\BIBentryALTinterwordstretchfactor{4}
\providecommand\BIBentryALTinterwordspacing{\spaceskip=\fontdimen2\font plus
\BIBentryALTinterwordstretchfactor\fontdimen3\font minus
  \fontdimen4\font\relax}
\providecommand\BIBforeignlanguage[2]{{%
\expandafter\ifx\csname l@#1\endcsname\relax
\typeout{** WARNING: IEEEtran.bst: No hyphenation pattern has been}%
\typeout{** loaded for the language `#1'. Using the pattern for}%
\typeout{** the default language instead.}%
\else
\language=\csname l@#1\endcsname
\fi
#2}}

\bibitem{cadena2016past}
C.~Cadena, L.~Carlone, H.~Carrillo, Y.~Latif, D.~Scaramuzza, J.~Neira, I.~Reid,
  and J.~J. Leonard, ``Past, present, and future of simultaneous localization
  and mapping: Toward the robust-perception age,'' \emph{IEEE Transactions on
  robotics}, vol.~32, no.~6, pp. 1309--1332, 2016.

\bibitem{tron2016survey}
R.~Tron, X.~Zhou, and K.~Daniilidis, ``A survey on rotation optimization in
  structure from motion,'' in \emph{Proceedings of the IEEE Conference on
  Computer Vision and Pattern Recognition Workshops}, 2016, pp. 77--85.

\bibitem{esquivel2007calibration}
S.~Esquivel, F.~Woelk, and R.~Koch, ``Calibration of a multi-camera rig from
  non-overlapping views,'' in \emph{Joint Pattern Recognition Symposium}, 2007,
  pp. 82--91.

\bibitem{agarwal2010bundle}
S.~Agarwal, N.~Snavely, S.~M. Seitz, and R.~Szeliski, ``Bundle adjustment in
  the large,'' in \emph{Proceedings of the European conference on computer
  vision}, 2010, pp. 29--42.

\bibitem{robotics9030051}
G.~Grisetti, T.~Guadagnino, I.~Aloise, M.~Colosi, B.~Della~Corte, and
  D.~Schlegel, ``Least squares optimization: From theory to practice,''
  \emph{Robotics}, vol.~9, no.~3, 2020.

\bibitem{g2o2011kummerle}
R.~{Kümmerle}, G.~{Grisetti}, H.~{Strasdat}, K.~{Konolige}, and W.~{Burgard},
  ``G2o: A general framework for graph optimization,'' in \emph{Proceedings of
  the IEEE International Conference on Robotics and Automation}, 2011, pp.
  3607--3613.

\bibitem{SAM2006dellaert}
F.~Dellaert and M.~Kaess, ``Square root sam: Simultaneous localization and
  mapping via square root information smoothing,'' \emph{The International
  Journal of Robotics Research}, vol.~25, no.~12, pp. 1181--1203, 2006.

\bibitem{chordalOriginal2007martinec}
D.~{Martinec} and T.~{Pajdla}, ``Robust rotation and translation estimation in
  multiview reconstruction,'' in \emph{Proceedings of the IEEE Conference on
  Computer Vision and Pattern Recognition}, 2007, pp. 1--8.

\bibitem{chordalAndInitialization2015carlone}
L.~{Carlone}, R.~{Tron}, K.~{Daniilidis}, and F.~{Dellaert}, ``Initialization
  techniques for 3d slam: A survey on rotation estimation and its use in pose
  graph optimization,'' in \emph{Proceedings of the IEEE International
  Conference on Robotics and Automation}, 2015, pp. 4597--4604.

\bibitem{hogman2010grisetti}
G.~{Grisetti}, R.~{Kümmerle}, C.~{Stachniss}, U.~{Frese}, and C.~{Hertzberg},
  ``Hierarchical optimization on manifolds for online 2d and 3d mapping,'' in
  \emph{Proceedings of the IEEE International Conference on Robotics and
  Automation}, 2010, pp. 273--278.

\bibitem{condensed2012grisetti}
G.~{Grisetti}, R.~{Kümmerle}, and K.~{Ni}, ``Robust optimization of factor
  graphs by using condensed measurements,'' in \emph{Proceedings of the
  IEEE/RSJ International Conference on Intelligent Robots and Systems}, 2012,
  pp. 581--588.

\bibitem{ni2007tectonic}
K.~Ni, D.~Steedly, and F.~Dellaert, ``Tectonic sam: Exact, out-of-core,
  submap-based slam,'' in \emph{Proceedings of the IEEE International
  Conference on Robotics and Automation}, 2007, pp. 1678--1685.

\bibitem{ni2010multi}
K.~Ni and F.~Dellaert, ``Multi-level submap based slam using nested
  dissection,'' in \emph{Proceedings of the IEEE/RSJ International Conference
  on Intelligent Robots and Systems}, 2010, pp. 2558--2565.

\bibitem{carlone2014angular}
L.~Carlone and A.~Censi, ``From angular manifolds to the integer lattice:
  Guaranteed orientation estimation with application to pose graph
  optimization,'' \emph{IEEE Transactions on Robotics}, vol.~30, no.~2, pp.
  475--492, 2014.

\bibitem{lu1997globally}
F.~Lu and E.~Milios, ``Globally consistent range scan alignment for environment
  mapping,'' \emph{Autonomous robots}, vol.~4, no.~4, pp. 333--349, 1997.

\bibitem{olson2006fast}
E.~Olson, J.~Leonard, and S.~Teller, ``Fast iterative alignment of pose graphs
  with poor initial estimates,'' in \emph{Proceedings of the IEEE International
  Conference on Robotics and Automation}, 2006, pp. 2262--2269.

\bibitem{grisetti2009nonlinear}
G.~Grisetti, C.~Stachniss, and W.~Burgard, ``Nonlinear constraint network
  optimization for efficient map learning,'' \emph{IEEE Transactions on
  Intelligent Transportation Systems}, vol.~10, no.~3, pp. 428--439, 2009.

\bibitem{davis2016sparsematrix}
T.~A. Davis, S.~Rajamanickam, and W.~M. Sid-Lakhdar, ``A survey of direct
  methods for sparse linear systems,'' \emph{Acta Numerica}, vol.~25, p.
  383–566, 2016.

\bibitem{isam2008kaess}
M.~{Kaess}, A.~{Ranganathan}, and F.~{Dellaert}, ``isam: Incremental smoothing
  and mapping,'' \emph{IEEE Transactions on Robotics}, vol.~24, no.~6, pp.
  1365--1378, 2008.

\bibitem{kaess2012isam2}
M.~Kaess, H.~Johannsson, R.~Roberts, V.~Ila, J.~J. Leonard, and F.~Dellaert,
  ``isam2: Incremental smoothing and mapping using the bayes tree,'' \emph{The
  International Journal of Robotics Research}, vol.~31, no.~2, pp. 216--235,
  2012.

\bibitem{ni2007out}
K.~Ni, D.~Steedly, and F.~Dellaert, ``Out-of-core bundle adjustment for
  large-scale 3d reconstruction,'' in \emph{Proceedings of the IEEE
  International Conference on Computer Vision}, 2007, pp. 1--8.

\bibitem{konolige2010efficient}
K.~Konolige, G.~Grisetti, R.~Kümmerle, W.~Burgard, B.~Limketkai, and
  R.~Vincent, ``Efficient sparse pose adjustment for 2d mapping,'' in
  \emph{Proceedings of the IEEE/RSJ International Conference on Intelligent
  Robots and Systems}, 2010, pp. 22--29.

\bibitem{hu2013towards}
G.~Hu, K.~Khosoussi, and S.~Huang, ``Towards a reliable slam back-end,'' in
  \emph{Proceedings of the IEEE/RSJ International Conference on Intelligent
  Robots and Systems}, 2013, pp. 37--43.

\bibitem{khosoussi2016cagate}
K.~Khosoussi, S.~Huang, and G.~Dissanayake, ``A sparse separable slam
  back-end,'' \emph{IEEE Transactions on Robotics}, vol.~32, no.~6, pp.
  1536--1549, 2016.

\bibitem{sharp2004multiview}
G.~C. Sharp, S.~W. Lee, and D.~K. Wehe, ``Multiview registration of 3d scenes
  by minimizing error between coordinate frames,'' \emph{IEEE Transactions on
  Pattern Analysis and Machine Intelligence}, vol.~26, no.~8, pp. 1037--1050,
  2004.

\bibitem{govindu2001combining}
V.~Govindu, ``Combining two-view constraints for motion estimation,'' in
  \emph{Proceedings of the IEEE Computer Society Conference on Computer Vision
  and Pattern Recognition}, 2001, pp. II--II.

\bibitem{sesync2019rosen}
D.~M. Rosen, L.~Carlone, A.~S. Bandeira, and J.~J. Leonard, ``Se-sync: A
  certifiably correct algorithm for synchronization over the special euclidean
  group,'' \emph{The International Journal of Robotics Research}, vol.~38, no.
  2-3, pp. 95--125, 2019.

\bibitem{Dellaert20eccv-shonan}
F.~Dellaert, D.~Rosen, J.~Wu, R.~Mahony, and L.~Carlone, ``Shonan rotation
  averaging: Global optimality by surfing {$SO(p)^n$},'' 2020.

\bibitem{bai2021cycle}
F.~{Bai}, T.~{Vidal-Calleja}, and G.~{Grisetti}, ``Sparse pose graph
  optimization in cycle space,'' \emph{IEEE Transactions on Robotics}, pp.
  1--20, 2021.

\bibitem{chordal2020aloise}
I.~{Aloise} and G.~{Grisetti}, ``Chordal based error function for 3-d
  pose-graph optimization,'' \emph{IEEE Robotics and Automation Letters},
  vol.~5, no.~1, pp. 274--281, 2020.

\bibitem{graphSLAM2010grisetti}
G.~{Grisetti}, R.~{Kümmerle}, C.~{Stachniss}, and W.~{Burgard}, ``A tutorial
  on graph-based slam,'' \emph{IEEE Intelligent Transportation Systems
  Magazine}, vol.~2, no.~4, pp. 31--43, 2010.

\bibitem{russel2020modern}
S.~J. Russell and P.~Norvig, \emph{Artificial Intelligence: {A} Modern Approach
  (4th Edition)}.\hskip 1em plus 0.5em minus 0.4em\relax Person, 2020.

\bibitem{kaess2009marginal}
M.~Kaess and F.~Dellaert, ``Covariance recovery from a square root information
  matrix for data association,'' \emph{Robotics and Autonomous Systems},
  vol.~57, no.~12, pp. 1198--1210, 2009.

\bibitem{powell1970new}
M.~J. Powell, ``A new algorithm for unconstrained optimization,'' in
  \emph{Nonlinear programming}, 1970, pp. 31--65.

\end{thebibliography}
	
\end{document}